\pdfoutput=1

\documentclass[11pt]{article}

\usepackage[]{ACL2023}

\usepackage{times}
\usepackage{latexsym}

\usepackage[T1]{fontenc}

\usepackage[utf8]{inputenc}

\usepackage{microtype}

\usepackage{inconsolata}

\usepackage{graphicx}
\usepackage{bm}
\usepackage{amsmath}
\usepackage{amsfonts}
\usepackage{bbm}

\usepackage{helvet}
\usepackage{courier}
\usepackage{multirow}
\usepackage{bm}
\usepackage{amsmath,amssymb}
\usepackage{mdwlist}

\usepackage{relsize}
\usepackage{amsfonts}
\usepackage{url}
\usepackage{graphicx}
\usepackage{subfigure}
\usepackage{caption}
\usepackage{hhline}
\usepackage{color}
\usepackage{colortbl}
\usepackage{booktabs}
\usepackage{tabularx}
\usepackage{makecell}
\usepackage[linesnumbered,boxed,ruled,commentsnumbered]{algorithm2e}
\usepackage{lipsum}
\title{RE-Matching: A Fine-Grained Semantic Matching Method for \\Zero-Shot Relation Extraction}

\author{Jun Zhao$^{1}$\footnotemark[1],\ \ Wenyu Zhan$^{1}$\footnotemark[1],\ \ Xin Zhao$^{1}$,\ \ Qi Zhang$^{1}$\footnotemark[2],\ \ Tao Gui$^{2}$\footnotemark[2], \\ \textbf{Zhongyu Wei}$^{3}$\textbf{,} \ \ \textbf{Junzhe Wang}$^1$\textbf{,}\ \ \textbf{Minlong Peng}$^4$\textbf{,}\ \ \textbf{Mingming Sun}$^4$\\
  $^1$School of Computer Science, Fudan University\\
  $^2$Institute of Modern Languages and Linguistics, Fudan University\\
  $^3$School of Data Science, Fudan University\\
  $^4$Cognitive Computing Lab Baidu Research\\
  \texttt{\{zhaoj19,qz,tgui\}@fudan.edu.cn, wyzhan21@m.fudan.edu.cn}}

\begin{document}
\maketitle
\renewcommand{\thefootnote}{\fnsymbol{footnote}}
\footnotetext[1]{Equal Contributions.}
\footnotetext[2]{Corresponding authors.}
\begin{abstract}
Semantic matching is a mainstream paradigm of zero-shot relation extraction, which matches a given input with a corresponding label description. 
The entities in the input should exactly match their hypernyms in the description, while the irrelevant contexts should be ignored when matching.
However, general matching methods lack explicit modeling of the above matching pattern. 
In this work, we propose a fine-grained semantic matching method tailored for zero-shot relation extraction. Following the above matching pattern, we decompose the sentence-level similarity score into entity and context matching scores. Due to the lack of explicit annotations of the redundant components, we design a feature distillation module to adaptively identify the relation-irrelevant features and reduce their negative impact on context matching.
Experimental results show that our method achieves higher matching $F_1$ score and has an inference speed 10 times faster, when compared with the state-of-the-art methods.
\end{abstract}

\section{Introduction}
Relation extraction (RE) is a fundamental task of natural language processing (NLP), which aims to extract the relations between entities in unstructured text. Benefitting from high-quality labeled data, neural relation extraction has achieved superior performance \cite{han-etal-2020-data,wu2019enriching,zhao-etal-2021-relation}.
However, it is expensive and even impractical to endlessly label data for a fast-growing number of new relations.

    \begin{figure}[t]
        \includegraphics[width=\columnwidth]{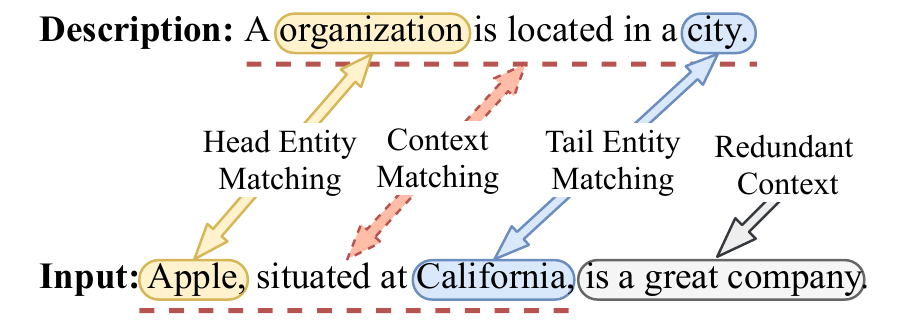}
        \caption{An example of the matching pattern of relational data. The input sentence contains a given entity pair, which should match the corresponding hypernyms (usually the entity type). The context describes the relations between entities, containing relation-irrelevant redundant information, which should be ignored when matching.}
        \label{fig:intro}
    \end{figure}
In order to deal with the emerging new relations that lack labeled data, zero-shot relation extraction (ZeroRE) has recently attracted more attention. 
\citet{levy-etal-2017-zero} frame the ZeroRE as a slot-filling task solved in a question-answering way. Each relation is associated with a few question templates. However, the templates are expensive and time-consuming to build \cite{chen-li-2021-zs}. \citet{obamuyide-vlachos-2018-zero} simplify the templates to readily available relational descriptions and reformulate ZeroRE as a semantic matching task. 
Recently, pretrained model based ZeroRE methods have achieved great success. \textit{Siamese scheme} and \textit{full encoding scheme} are two mainstream methods for matching semantics. The siamese scheme separately encodes the input and description \cite{chen-li-2021-zs}. Therefore, the encoded representations of descriptions can be both stored and reused for each input, resulting in a fast inference. However, insufficient interaction during encoding also limits the matching performance. By contrast, the full encoding scheme performs self-attention over the pair to enrich interaction \cite{sainz-etal-2021-label}, although the performance increase comes with a computational overhead. (For $m$ inputs and $n$ descriptions, the siamese scheme requires $m+n$ encodings, while the number is $m\times n$ for full encoding). An approach that combines the advantages of both can be attractive.

Unlike ordinary sentence pairs, relational data has a unique matching pattern, which is not explicitly considered by general matching methods. As shown in fig. \ref{fig:intro}, the entities in the input should exactly match their hypernyms in the description (e.g., Apple and organization). Meanwhile, not all contextual words contribute equally to the relation semantics. For example, the clause ``is a great company'' is only used to modify Apple instead of expressing the relationship between Apple and California. Such redundant components should be ignored when matching. Due to the lack of explicit annotations to the redundant components, it is non-trivial for the model to learn to identify them.

In this work, we propose a fine-grained semantic matching method that improves both the accuracy and speed over the current state-of-the-art. Specifically, we decouple encoding and matching into two modules. While the encoding module follows a siamese scheme for efficiency, the matching module is responsible for the fine-grained interaction.
Following the matching pattern of relational data, the sentence-level similarity score is decomposed into two: entity matching and context matching scores. 
To deal with the redundant components without explicit annotations, we design a feature distillation module. 
Context features that maximize a classification loss are identified as relation-irrelevant features.
Then, the context representations are projected into the orthogonal space of the features to improve context matching.
Experimental results show that this method outperforms state-of-the-art (SOTA) methods for ZeroRE, in terms of both accuracy and speed. Our codes are publicly available\footnote{https://github.com/zweny/RE-Matching}.

The main contributions are three-fold: (1) We propose a fine-grained semantic matching method for ZeroRE, which explicitly models the matching pattern of relational data; (2) We propose a context distillation method, which can reduce the negative impact of irrelevant components on context matching; (3) Experimental results show that our method achieves SOTA matching $F_1$ score together with an inference speed 10 times faster.



\section{Related Works}
Text semantic matching aims to predict a matching score that reflects the semantic similarity between a given pair of text sequences. The approaches used to calculate the matching score roughly fall into one of two groups. The first group uses a siamese scheme, which separately maps the pair of text sequences into a common feature space wherein a dot product, cosine, or parameterized non-linearity is used to measure the similarity \cite{10.1145/1645953.1645979,10.1145/2505515.2505665,10.1145/3269206.3271800,wu2018starspace,chen-li-2021-zs}. The second group adopts a full encoding scheme, in which the concatenation of the text pair serves as a new input to a nonlinear matching function. Neural networks with different inductive biases are used to instantiate the matching function \cite{wu-etal-2017-sequential,10.1145/3209978.3210011,zhang-etal-2018-modeling}. 
More recently, large-scale pretrained language models such as BERT \cite{devlin-etal-2019-bert}, RoBERTa \cite{DBLP:journals/corr/abs-1907-11692} 
are introduced to yield richer interactions between the text pair, to improve semantic matching.

Unlike the general matching methods, the proposed method is designed to be specifically used for relational data. By explicitly modeling the matching pattern, our method achieves SOTA performance while decreasing the computational cost by an order of magnitude compared with a full encoding scheme (as the more candidate relations exist, the more cost decreases).

\begin{figure*}[t]
    \centering
        \includegraphics[width=\linewidth]{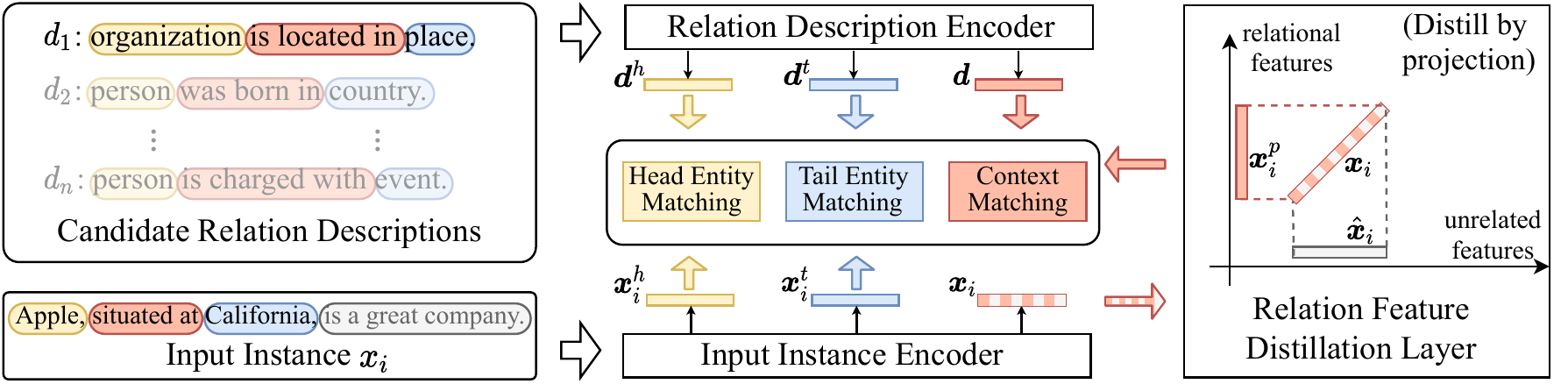}
        \caption{Overview of the proposed RE-Matching method. The input instance and candidate relation descriptions (left side) are separately encoded for efficiency. To model the matching pattern of relational data, we calculate similarity by entity and context matching (middle). In addition, we design a distillation module to reduce the impact of irrelevant components (the gray part in the input) on context matching (right side).}
        \label{fig:model}
    \end{figure*} 
\section{Approach}
In this work, we propose a fine-grained semantic matching method tailored for relational data. To facilitate inference efficiency, we adopt a siamese scheme to separately encode the input and the candidate relation descriptions. To explicitly model the matching pattern of relational data, we decompose the matching score into entities and contexts matching scores. In addition, we design a context distillation module to reduce the negative impact of irrelevant components in the input on context matching.

\subsection{Task Formulation and Method Overview}

\noindent\textbf{Task Formulation:}
In Zero-shot relation extraction (ZeroRE), the goal is to learn from the seen relations $\mathcal{R}_s=\{r_1^s,r_2^s,...,r_n^s\}$ and generalize to the unseen relations $\mathcal{R}_u=\{r_1^u,r_2^u,...,r_m^u\}$. Such two sets are disjoint, i.e., $\mathcal{R}_s\cap\mathcal{R}_u=\varnothing$ and only the samples of the seen relations $\mathcal{R}_s$ are available at the training phase. Following \citet{chen-li-2021-zs}, we formulate ZeroRE as a semantic matching problem. Specifically, given the training set $\mathcal{D}=\{(x_i,e_{i1},e_{i2},y_i,d_i)|i=1,..,N\}$ with $N$ samples, consisting of input instance $x_i$, target entity pair $e_{i1}$ and $e_{i2}$, relation $y_i\in \mathcal{R}_s$ and its corresponding relation description $d_i$, we optimize a matching model $\mathcal{M}(x,e_1,e_2,d)\rightarrow s\in\mathbb{R}$, where the score $s$ measures the semantic similarity between the input instance $x$ and the relation description $d$. When testing, we transfer the matching model $\mathcal{M}$ to extract unseen relations in $\mathcal{R}_u$. Specifically, given a sample $(x_j, e_{j1},e_{j2})$ expressing an unseen relation in $\mathcal{R}_u$, we make prediction by finding the relation $\hat{y}_j$ whose description has the highest similarity score with the input sample.

\noindent\textbf{Method Overview:} We approach the problem with a fine-grained semantic matching method tailored for relational data. As illustrated in fig. \ref{fig:model}, we decouple encoding and matching into two modules, explicitly modeling the matching pattern of relational data while ensuring efficient inference.

The encoding module is designed to extract both entity and contextual relational information from an input instance $x_i$ and a candidate description $d\in\{d_{r^s_i}|i=1,..,n\}$, which lays the groundwork for fine-grained matching. To enrich the insufficient entity information in $d$, we separate the hypernyms (i.e. entity types) in $d$ and design several methods to automatically expand them into a complete head (tail) entity description $d^{h}(d^{t})$. We adopt a fixed Sentence-BERT \cite{2019Sentence} to encode the $d$, $d^{h}$, $d^{t}$ to their representations $\bm{d}$, $\bm{d}^{h}$, $\bm{d}^{t}\in\mathcal{R}^n$, respectively. For input $x_i$, we use a trainable BERT \cite{devlin-etal-2019-bert} to encode its context and entity representation $\bm{x}_i$, $\bm{x}_i^{h}$, $\bm{x}_i^{t}\in\mathcal{R}^n$. Because descriptions and instances are separately encoded, the computational complexity is reduced from $\mathcal{O}(mn)$ to $\mathcal{O}(m+n)$, compared with the full encoding scheme ($m,n$ represent the number of candidate descriptions and input instances, respectively).

The matching module is responsible for the interaction between input $x_i$ and description $d$. The score of entity matching is directly calculated with cosine similarity $\cos(\bm{x}_i^h, \bm{d}^h)$ and $\cos(\bm{x}_i^t, \bm{d}^t)$. To reduce the impact of redundant information in $x_i$, the context representation $\bm{x}_i$ is fed into the distillation module, where $\bm{x}_i$ is projected into the orthogonal space of irrelevant features to obtain the refined representation $\bm{x}_i^p$. The score of context matching is $\cos(\bm{x}_i^p, \bm{d})$. Finally, the overall matching score is the sum of entity and context matching scores.

\subsection{Input-Description Encoding Module}
The encoding module aims to encode the entity and context information in the input and label description into fixed-length representations for subsequent fine-grained matching.
\subsubsection{Relation Description Encoder}
Each candidate relation corresponds to a natural language label description $d\in\{d_{r^s_i}|i=1,..,n\}$. For example, the relation \textit{headquartered\_in} corresponds to the description ``\textit{the headquarters of an organization is located in a place}'', and its encoded representation $\bm{d}$ can be used for contextual matching. However, how an entity description is constructed (based on $d$), is important for a high-quality entity representation. In this subsection, we explore different ways to automatically construct and enrich entity description as follows:

\noindent\textbf{Keyword:} The entity hypernym (i.e. entity type) in $d$ is directly used as the entity description. Take $\textit{headquartered\_in}$ as an example, $d^{h}$ is ``\textit{organization}'' and $d^{t}$ is ``\textit{place}''.
    
\noindent\textbf{Synonyms:} To further enrich the entity information, we use the words that mean exactly or nearly the same as the original hypernym extracted using Wikidata\footnote{https://www.wikidata.org/} and Thesaurus \footnote{https://www.thesaurus.com/}. Then, $d^{h}$ becomes ``\textit{organization, institution, company}''.
    
\noindent\textbf{Rule-based Template Filling:} Inspired by prompt learning \cite{DBLP:journals/corr/abs-2107-13586}, a fluent and complete entity description may stimulate the pretrained model to output a high-quality entity representation. The synonym-extended hypernym sequence is filled into a template with some slots (i.e. \textit{the head/tail entity types including [S], [S], ...}), and $d^{h}$ becomes ``\textit{the head entity types including organization, institution, company}''. Empirical results show that expanding all candidate descriptions with the above template works well. The customized template design is not the focus of this paper, so we leave it to future work.

Following \cite{chen-li-2021-zs}, we adopt a fixed sentence-BERT as the implementation of the relation description encoder $\bm{f}(\cdot)$, which encodes the above descriptions into fixed-length representations, that is $\bm{f}(d)=\bm{d}\in \mathbb{R}^d$, $\bm{f}(d^{h})=\bm{d}^{h}\in \mathbb{R}^d$, $\bm{f}(d^{t})=\bm{d}^{t}\in \mathbb{R}^d$.


\subsubsection{Input Instance Encoder}
\label{sec:IIE}
Given an input instance $x_i=\{w_1,w_2,...,w_n\}$, in which four reserved special tokens $[E_h],[\backslash E_h], [E_t],[\backslash E_t]$ are inserted to mark the beginning and end of the head entity $e_{i1}$ and tail entity $e_{i2}$ respectively. We obtain entity representation $\bm{x}_i^h$ and $\bm{x}_i^h$ by MaxPool the corresponding hidden states of entity tokens. Following \cite{baldini-soares-etal-2019-matching}, the context representation $\bm{x}_i$ is obtained by concatenating the hidden states of special token $[E_h]$, $[E_t]$.
        \begin{gather}
            \bm{h}_1,...,\bm{h}_n=\text{BERT}(w_1,...,w_n)\\
            \bm{x}_i^{h}=\text{MaxPool}(\bm{h}_{b_h},...,\bm{h}_{e_h})\\
            \bm{x}_i^{t}=\text{MaxPool}(\bm{h}_{b_t},...,\bm{h}_{e_t})\\
            \bm{x}_{i}=\phi(\left \langle\bm{h}_{[E_h]}|\bm{h}_{[E_t]}\right \rangle),
            \label{equ:cat}
        \end{gather}
where $\left \langle\cdot | \cdot \right \rangle$ denotes the concatenation operator. $b_h, e_h, b_t, e_t$ denote the beginning and end position indexes of the head and tail entities respectively. $\bm{h}_{[E_h]}$ and $\bm{h}_{[E_t]}$ represent the hidden states of $[E_h]$ and $[E_t]$ respectively. Their corresponding position indexes are $b_h-1$ and $b_t-1$. $\phi(\cdot)$ denotes a linear layer with tanh activation, converting the dimension of $\left \langle\bm{h}_{[E_h]}|\bm{h}_{[E_t]}\right \rangle$ from $2n$ back to $n$.

\subsection{Contextual Relation Feature Distillation}
Due to the lack of explicit annotations to the relation-irrelevant components, it is non-trivial for the model to learn to identify them. This section introduces how to reduce the negative impact of the redundant components on context matching.

\subsubsection{Relation-Irrelevant Feature Aggregator}
Given the output $\bm{h}_1,...,\bm{h}_n$ of the input instance encoder, we aggregate the relation-irrelevant features through a trainable query code $\bm{q}\in \mathbb{R}^d$ as follows:
        \begin{gather}
            (\alpha_1,...,\alpha_n)=\text{Softmax}(\bm{q}\cdot\bm{h}_1,...,\bm{q}\cdot\bm{h}_n)\\
            \bm{x}^*_i=\sum_{j=1}^n\alpha_j\cdot\bm{h}_j,
            \label{equ:unrelated}
        \end{gather}
This leads to a immediate question, \textbf{how do we make query code $\bm{q}$ select relation-irrelevant features from context?} Intuitively, it is impossible for a relational classifier to discriminate the relations of input instances based on relation-irrelevant features. Therefore, we introduce a Gradient Reverse Layer (GRL) \cite{pmlr-v37-ganin15} and optimize $\bm{q}$ by fooling the relational classifier.
        \begin{gather}
            prob_i=\text{Softmax}(\text{GRL}(\bm{x}^*_i)\cdot W+b)\\
            \mathcal{L}_{ce,i}=\text{CrossEntropy}(y_i,prob_i),
        \end{gather}
where $W$ and $b$ are the weights and biases of the relation classifier. $\bm{x}^*_i$ goes through a GRL layer before being fed into the classifier. GRL does not affect forward propagation but changes the gradient sign during backpropagation by multiplying $-\lambda$. That is, as the training proceeds, the classifier is optimized by gradient descent to reduce $\mathcal{L}_{ce,i}$, while the query code $\bm{q}$ is optimized by gradient ascent to increase $\mathcal{L}_{ce,i}$, until no relational features are included in $\bm{x}^*_i$. The effectiveness of GRL has been verified in many literatures in the past \cite{10.5555/2946645.2946704,10.1609/aaai.v33i01.33015773}.

\subsubsection{Relation Feature Distillation Layer}
The distillation module aims to reduce the negative impact of relation-irrelevant components on its representation and thus improving context matching. Given a context representation $\bm{x}_i$ (refer to eq.\ref{equ:cat}), as well as relation-irrelevant features $\bm{x}_i^*$ (refer to eq.\ref{equ:unrelated}), we achieve the above goal by projecting $\bm{x}_i$ to the orthogonal space of $\bm{x}_i^*$. Specifically, we first project $\bm{x}_i$ to the direction of $\bm{x}_i^*$ to find the relation-irrelevant features $\hat{\bm{x}}_i$ mixed in $\bm{x}_i$ as follows:
        \begin{gather}
            \hat{\bm{x}}_i=\text{Proj}(\bm{x}_i, \bm{x}_i^*)\label{equ:proj}\\
            Proj(\bm{a},\bm{b}) = \frac{\bm{a}\cdot \bm{b}}{|\bm{b}|}\cdot\frac{\bm{b}}{|\bm{b}|},
        \end{gather}
where $\text{Proj}(\cdot, \cdot)$ denotes the projection operator and $\bm{a}$, $\bm{b}$ are the input vectors of $\text{Proj}$. Then, we obtain the refined context representation $\bm{x}_i^p$ by removing $\hat{\bm{x}}_i$ from $\bm{x}_i$ as follows:
        \begin{gather}
            \bm{x}_i^p=\bm{x}_i-\hat{\bm{x}}_i.
            \label{equ:rev}
        \end{gather}
Essentially, eqs. \ref{equ:proj}-\ref{equ:rev} project $\bm{x}_i$ in the orthogonal direction of $\bm{x}_i^*$. The above process is illustrated in the right side of fig. \ref{fig:model}.
\subsection{Fine-Grained Matching and Training}
Following the matching pattern of relational data, we decompose sentence-level matching into entity matching and context matching. For an input $x_i$ and a candidate relation description $d\in\{d_{r^s_i}|i=1,..,n\}$, the entity and context information are encoded into fixed-length representations ($\bm{x}_i^{h}$, $\bm{x}_i^{t}$, $\bm{x}_i^{p}$), and ($\bm{d}^{h}$, $\bm{d}^{t}$, $\bm{d}$), respectively. The matching score between $x_i$ and $d$ is the sum of entity and context matching scores as follows:
    \begin{align}
        s(x_i,d)=&\alpha\cdot\cos(\bm{x}_i^{h},\bm{d}^{h})+\alpha\cdot  \cos(\bm{x}_i^{t},\bm{d}^{t})\nonumber\\
        &+(1-2\cdot\alpha)\cdot \cos(\bm{x}_i^p,\bm{d}),
    \end{align}
where $\alpha$ is a hyper-parameter and $\cos(\cdot, \cdot)$ denotes the cosine operator. In order to optimize the above matching model and avoid over-fitting, we use margin loss as the objective function.
    \begin{gather}
        \delta_i=s(x_i,d_{y_i})-\max_{j\neq y_i}(s(x_i,d_j))\\
        \mathcal{L}_{m,i}=\max(0, \gamma-\delta_i),
    \end{gather}
where $\gamma>0$ is a hyper-parameter, meaning that the matching score of a positive pair must be higher than the closest negative pair. The overall training objective is as follows:
\begin{gather}
        \mathcal{L}=\frac{1}{N}\sum_{i=1}^{N}(\mathcal{L}_{ce,i}+\mathcal{L}_{m,i}),
    \end{gather}
where $N$ is the batch size. When testing, the learned model is transferred to recognize the unseen relations in $\mathcal{R}_u$, by a match between the input and the descriptions of the unseen relations.

\section{Experimental Setup}

\subsection{Datasets}
\label{sec:dataset}
\textbf{FewRel} \cite{han-etal-2018-fewrel} is a few-shot relation classification dataset collected from Wikipedia and further hand-annotated by crowd workers, which contains 80 relations and 700 sentences in each type of relation. \textbf{Wiki-ZSL} \cite{chen-li-2021-zs} is derived from Wikidata Knowledge Base and consists of 93,383 sentences on 113 relation types. Compared with the FewRel dataset, Wiki-ZSL has more abundant relational information but inevitably has more noise in raw data since it is generated by distant supervision.

Following \citet{chia-etal-2022-relationprompt}, we randomly select 5 relations for validation set, $m\in\{5,10,15\}$ relations as unseen relations for testing set, and consider the remaining relations as seen relations for training set. Meanwhile, we randomly repeat the class selection process 5 times to ensure the reliability of the experiment results. We report the average results across different selections.


        \begin{table*}
            \centering
            \resizebox{\linewidth}{!}{
            \begin{tabular}{cl lll lll}
            \toprule
            \multirow{2}{*}{\textbf{Unseen Labels}} & \multirow{2}{*}{\textbf{Method}} & \multicolumn{3}{c}{Wiki-ZSL} & \multicolumn{3}{c}{FewRel}\\
            \cline{3-8}
            & & Prec. & Rec. & $F_1$ & Prec. & Rec. & $F_1$\\
            \midrule
            \multirow{7}{*}{$m=5$}
            & R-BERT \cite{10.1145/3357384.3358119} & 39.22 & 43.27 & 41.15 & 42.19 & 48.61 & 45.17\\
            & ESIM \cite{levy-etal-2017-zero} & 48.58 & 47.74 & 48.16 & 56.27 & 58.44 & 57.33\\
            &ZS-BERT \cite{chen-li-2021-zs} & 71.54 & 72.39 & 71.96 & 76.96 & 78.86 & 77.90\\
            &PromptMatch  \cite{sainz-etal-2021-label} & 77.39 & 75.90 & 76.63 & 91.14 & 90.86 & 91.00\\
            &REPrompt(NoGen) \cite{chia-etal-2022-relationprompt}& 51.78 & 46.76 & 48.93 & 72.36 & 58.61 & 64.57\\
            &REPrompt \cite{chia-etal-2022-relationprompt}& 70.66 & 83.75 & 76.63 & 90.15 & 88.50 & 89.30\\
            &\cellcolor{gray!20}\textbf{RE-Matching}& \cellcolor{gray!20}$78.19$ & \cellcolor{gray!20}$78.41$ & \cellcolor{gray!20}$\textbf{78.30}$ & \cellcolor{gray!20}$92.82$ & \cellcolor{gray!20}$92.34$ & \cellcolor{gray!20}$\textbf{92.58}$\\
            \hline \hline
            \multirow{7}{*}{$m=10$}
            & R-BERT \cite{10.1145/3357384.3358119} & 26.18 & 29.69 & 27.82 & 25.52 & 33.02 & 28.20\\
            & ESIM \cite{levy-etal-2017-zero} & 44.12 & 45.46 & 44.78 & 42.89 & 44.17 & 43.52\\
            &ZS-BERT \cite{chen-li-2021-zs} & 60.51 & 60.98 & 60.74 & 56.92 & 57.59 & 57.25\\
            &PromptMatch \cite{sainz-etal-2021-label} & 71.86 & 71.14 & 71.50 & 83.05 & 82.55 & 82.80\\
            &REPrompt(NoGen) \cite{chia-etal-2022-relationprompt}& 54.87 & 36.52 & 43.80 & 66.47 & 48.28 & 55.61\\
            &REPrompt \cite{chia-etal-2022-relationprompt}& 68.51 & 74.76 & 71.50 & 80.33 & 79.62 & 79.96\\
            &\cellcolor{gray!20}\textbf{RE-Matching}& \cellcolor{gray!20}$74.39$ & \cellcolor{gray!20}$73.54$ & \cellcolor{gray!20}$\textbf{73.96}$ & \cellcolor{gray!20}$83.21$ & \cellcolor{gray!20}$82.64$ & \cellcolor{gray!20}$\textbf{82.93}$\\
            \hline \hline
            \multirow{7}{*}{$m=15$}
            & R-BERT \cite{10.1145/3357384.3358119} & 17.31 & 18.82 & 18.03 & 16.95 & 19.37 & 18.08\\
            & ESIM \cite{levy-etal-2017-zero} & 27.31 & 29.62 & 28.42 & 29.15 & 31.59 & 30.32\\
            &ZS-BERT \cite{chen-li-2021-zs} & 34.12 & 34.38 & 34.25 & 35.54 & 38.19 & 36.82\\
            &PromptMatch \cite{sainz-etal-2021-label} & 62.13 & 61.76 & 61.95 & 72.83 & 72.10 & 72.46\\
            &REPrompt(NoGen) \cite{chia-etal-2022-relationprompt}& 54.45 & 29.43 & 37.45 & 66.49 & 40.05 & 49.38\\
            &REPrompt \cite{chia-etal-2022-relationprompt}& 63.69 & 67.93 & 65.74 & 74.33 & 72.51 & 73.40\\
            &\cellcolor{gray!20}\textbf{RE-Matching}& \cellcolor{gray!20}$67.31$ & \cellcolor{gray!20}$67.33$ & \cellcolor{gray!20}$\textbf{67.32}$ & \cellcolor{gray!20}$73.80$ & \cellcolor{gray!20}$73.52$ & \cellcolor{gray!20}$\textbf{73.66}$\\
            \bottomrule
            \end{tabular}
            }
            \caption{Main results on two relation extraction datasets. We report the average results of five runs and the improvement is significant (using a Wilcoxon signed-rank test; $p < 0.05)$.}
            \label{tab:main_res}
        \end{table*}    
        
\subsection{Compared Methods}
To evaluate the effectiveness of our method, we make comparisons with a classic supervised method and state-of-the-art matching-based ZeroRE methods. We also compare a recent competitive seq2seq-based ZeroRE method.

\noindent\textbf{R-BERT} \cite{10.1145/3357384.3358119}. A SOTA supervised RE method. Following \citet{chen-li-2021-zs}, we adapt it to zero-shot setting by using the sentence representation to perform nearest neighbor search and generate zero-shot prediction.

\noindent\textbf{ESIM} \cite{levy-etal-2017-zero}. A classical matching-based ZeroRE method, which uses Bi-LSTM to encode the input and label description.

\noindent\textbf{ZS-BERT} \cite{chen-li-2021-zs}. A SOTA siamese-based ZeroRE method, which adopts BERT as the encoder to separately encode the input and relation description. In addition to classification loss, a metric-based loss is used to optimize representation space to improve nearest neighbor search.

\noindent\textbf{PromptMatch} \cite{sainz-etal-2021-label}. A SOTA full encoding-based ZeroRE method, which adopts BERT to encode the concatenation of input pairs and model their fine-grained semantic interaction.

\noindent\textbf{REPrompt} \cite{chia-etal-2022-relationprompt}. This baseline is a competitive seq2seq-based ZeroRE method. It uses GPT-2 to generate pseudo data of these relations to finetune the model. We use NoGen to denote the results without data augmentation.
\subsection{Implementation Details}
We use \textit{Bert-base-uncased} as the input instance encoder, and we adopt a fixed sentence-Bert\cite{2019Sentence}  \textit{stsb-bert-base} as the relation description encoder.  We set AdamW\cite{loshchilov2017decoupled} as the optimizer, and $2e-6$ as the learning rate. Based on the validation set, we conduct hyper-parameter selection. $\alpha$ is selected among \{0.2, 0.33, 0.4\} and $\lambda$ is selected among \{0.1, 0.3, 0.5, 0.7\}. Finally, we set $\alpha=0.33$ and $\lambda=0.5$ for all datasets. The batch size is set to 128. All experiments are conducted using an NVIDIA GeForce RTX 3090.

\section{Results and Analysis}

\subsection{Main Results}

The results on Wiki-ZSL and FewRel datasets are reported in tab. \ref{tab:main_res}, which shows that the proposed method consistently outperforms previous SOTA methods when targeting at a different number of unseen relations. Specifically, simple classification loss only focuses on the discrimination between known relations, so the supervised method such as R-BERT fails on ZeroRE. Although ZS-BERT is designed for ZeroRE, the siamese scheme limits the word-level interaction between the input and relation description, leading to suboptimal performance. By contrast, our method compensates for the shortcoming by explicit modeling of fine-grained matching patterns of relational data, thereby outperforming ZS-BERT by a large margin. Although full encoding scheme such as PromptMatch can implicitly model fine-grained interactions through self-attention, the proposed method is still able to outperform it. One possible reason is that the relational matching pattern, as an inductive bias, alleviates the over-fitting to seen relations in the training set and thus our model has better generalization.
Compared with REPrompt, which is a seq2seq-based method, our method achieves better results without using pseudo data of new relations to fine-tune the model, showing the superiority of the method.

\subsection{Ablation Study}

To study the contribution of each component in the proposed method, we conduct ablation experiments on the two datasets and display the results in tab. \ref{tab:abalation}. The results show that the matching performance is declined if the context distillation module is removed (i.e., w/o Proj), indicating that the relation-irrelevant information in the context disturbs the match of relational data, and projection in the distillation module is effective for reducing this impact. It is worth noting that entity information plays an important role in relational data (see w/o Ent). Explicitly modeling the matching between entities and their hypernyms significantly improves the performance of the model. As two vital components, when both context distillation and entity matching are removed (i.e., w/o both), the matching degenerates into sentence-level matching and the performance will be seriously hurt.
        \begin{table}
            \centering
            \resizebox{\columnwidth}{!}{
            \begin{tabular}{cl lll}
            \toprule
            \textbf{Dataset} & \textbf{Method} & Prec. & Rec. & $F_1$\\
            \midrule
            \multirow{4}{*}{\textbf{Wiki-ZSL}}
            &w/o Proj. & $66.13$ & $67.18$ & $66.65$\\
            &w/o Ent. & $41.81$ & $40.46$ & $41.12$\\
            &w/o both & $36.34$ & $36.12$ & $36.23$\\
            &\textbf{Ours}& $67.31$ & $67.33$ & $\textbf{67.32}$\\
            \hline
            \multirow{4}{*}{\textbf{FewRel}}
            &w/o Proj. & $72.35$ & $71.24$ & $71.79$\\
            &w/o Ent. & $49.16$ & $41.51$ & $45.01$\\
            &w/o both & $37.20$ & $32.43$ & $34.65$\\
            &\textbf{Ours}& $73.80$ & $73.52$ & $\textbf{73.66}$\\
            \bottomrule
            \end{tabular}
            }
            \caption{Ablation study of our method $(m=15)$.}
            \label{tab:abalation}
        \end{table}    

\subsection{Efficiency Advantage}
        \begin{figure}[t]
            \includegraphics[width=\columnwidth]{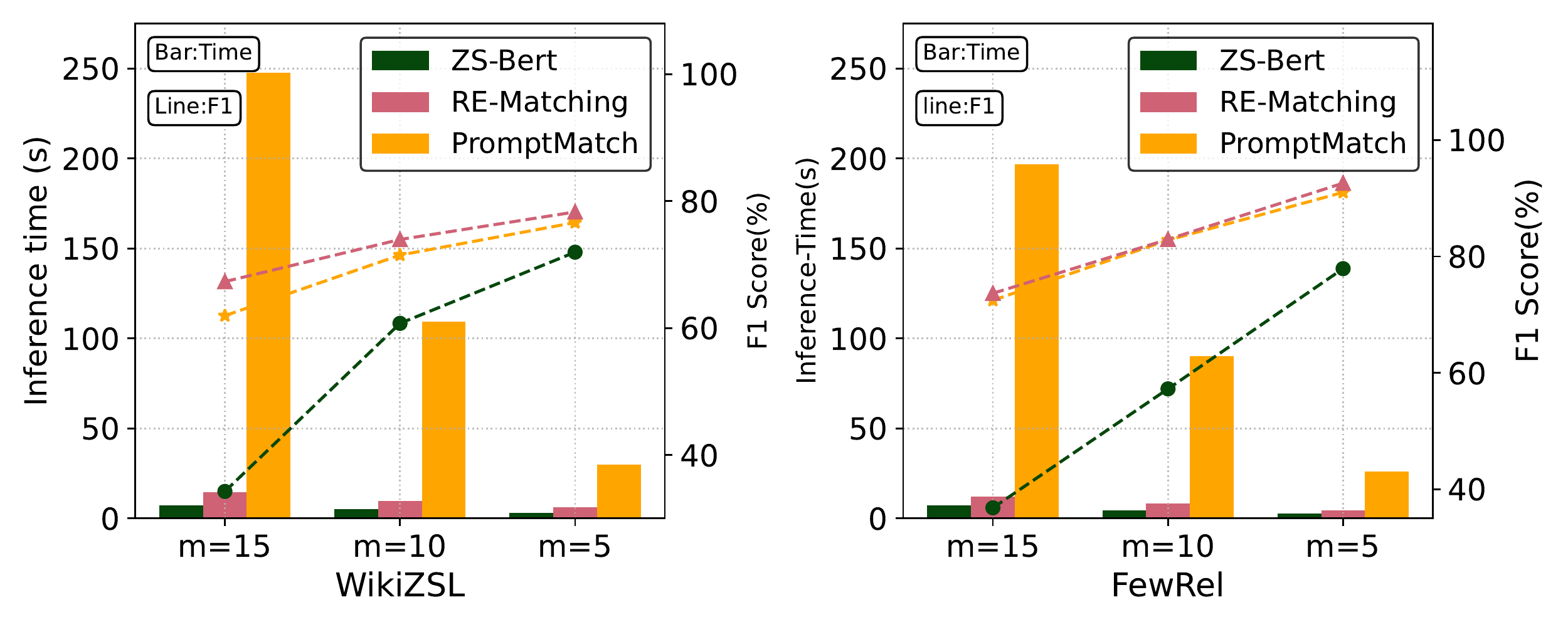}
            \caption{Comparison in terms of time consumption (Bars) and matching $F_1$ (Dotted lines). The fine-grained matching (pink) improves $F_1$ while maintaining the efficiency of the siamese scheme (i.e., ZS-BERT).}
            \label{fig:speed}
        \end{figure}
Fig. \ref{fig:speed} shows the time consumption and matching $F_1$ scores on Wiki-ZSL and FewRel datasets. 
Take FewRel as an example, each relation contains 700 testing inputs. The siamese scheme (ZS-Bert and our RE-Matching) separately encodes input and descriptions and the encoding is run ($700\cdot m+m$) times. By contrast, the full encoding scheme (PromptMatch) requires the concatenation of the text pair to be fed and the encoding is run ($700\cdot m^2$) times. Clearly, as the number of new relations $m$ increases, ZS-Bert and our RE-Matching have a significant efficiency advantage over PromptMatch that adopts a full encoding scheme. When $m=15$, the inference time can be reduced by more than 10 times. Although our method takes slightly more time than ZS-BERT, the fine-grained matching brings a significant $F_1$ improvement. As shown in tab. \ref{tab:main_res}, when $m=15$, our method improves the $F_1$ score by 33.07\% and 36.84\% on two datasets respectively, compared with ZS-BERT.

\subsection{Consistency over Various Encoders}
        \begin{table}
            \centering
            \resizebox{\columnwidth}{!}{
            \begin{tabular}{lcc}
            \toprule
            \multirow{2}{*}{\textbf{Model}} & \textbf{Wiki-ZSL} & \textbf{FewRel}\\
            &FullEncoding$\rightarrow$Ours (change)&FullEncoding$\rightarrow$Ours (change)\\
            \midrule
            BERT&$61.95\rightarrow67.32(\bm{5.37}\uparrow)$ & $72.46\rightarrow73.66(\bm{1.20}\uparrow)$\\
            RoBERTa&$62.58\rightarrow72.86(\bm{10.28}\uparrow)$ & $73.79\rightarrow73.82(0.03\uparrow)$\\
            DistilBERT&$57.05\rightarrow67.41(\bm{10.36}\uparrow)$ & $66.34\rightarrow69.35(\bm{3.01}\uparrow)$\\
            DistilRoBERTa&$57.46\rightarrow68.14(\bm{10.68}\uparrow)$ & $68.13\rightarrow70.73(\bm{2.60}\uparrow)$\\
            \bottomrule
            \end{tabular}
            }
            \caption{$F_1$ scores on two datasets. We compare the fine-grained matching with a full encoding scheme by varying the backbone to show consistency. Numbers in \textbf{bold} indicate whether the change is significant (using a Wilcoxon signed-rank test; $p < 0.05$).}
            \label{tab:plm}
        \end{table}    
In this section, we evaluate the effectiveness of our method by varying the selection of encoders. Tab. \ref{tab:plm} shows the comparison results between ours and the full encoding scheme (i.e., PromptMatch) when $m=15$. It can be observed that our method achieves consistent improvement.
PromptMatch is able to learn the word-level interaction by self-attention on the input-description pair. However, the data-driven optimization paradigm usually learns spurious correlations in data pairs, especially in noisy data.
By contrast, the matching pattern can be seen as a reasonable inductive bias. Modeling the pattern can reduce the overfitting risk. Therefore, our method consistently outperforms PromptMatch, especially on the noisy Wiki-ZSL (A distantly supervised dataset. See sec. \ref{sec:dataset}).
        \begin{figure*}[t]
            \includegraphics[width=\linewidth]{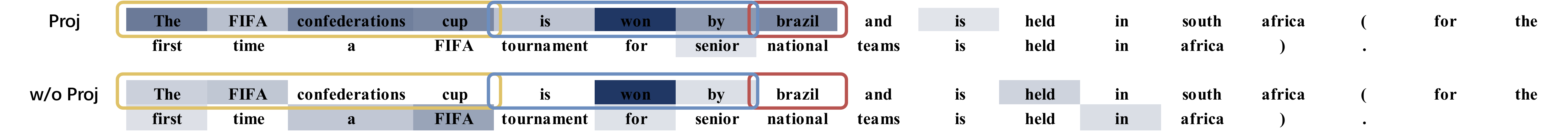}
            \caption{Visualization of the impact of words on context matching score. Darker colors indicate greater impact. The upper/lower sub-figure shows the results with/without projection respectively. The yellow and red boxes identify the head and tail entities respectively, while the blue box identifies the relational phrases [Relation: Winner].}
            \label{fig:case}
        \end{figure*}
        
\subsection{Error Analysis}
\noindent\textbf{What errors of baselines our method is able to correct}? (1) Our method reduces the negative impact of irrelevant components on context matching by projection. To intuitively show this, we use the attribution technique \cite{feng-etal-2018-pathologies} to find words that are highly influential to the context matching score. A visualization case is shown in fig. \ref{fig:case}. When using the projection, the model pays more attention to entities and relational phrases instead of irrelevant components (e.g., held, first, a) to make the prediction.
(2) The entity matching score can provide more information to distinguish confusing relations. Taking P59:constellation\_of\_celestial\_sphere as an example, its $F_1$ is only 0.123 without entity matching score. In this example, 79.71\% of the incorrect cases are recognized as P361:part\_of, due to the fact that the descriptions of P361 and P59 are similar (a constellation is described as part of a celestial sphere). With entity matching, the type of head and tail entities is constrained and the $F_1$ is improved from 0.123 to 0.950.

\noindent\textbf{What errors need to be addressed in further studies}? 
Taking P937:place\_of\_work and P19:place\_of\_birth as examples, the entity pair type of both relations are person-location. Therefore, explicitly modeling entity matching does not lead to further improvement, when compared with the baselines. In addition, some abstract relations are difficult to accurately recognize. Take P460:said\_to\_be\_the\_same as an example. Such abstract relations do not have explicit entity types, and it is difficult to give a high-quality relation description. Therefore, the $F_1$ score of P460 is only 0.03.


\subsection{Hyper-parameter Analysis}
        \begin{figure}[t]
            \includegraphics[width=\columnwidth]{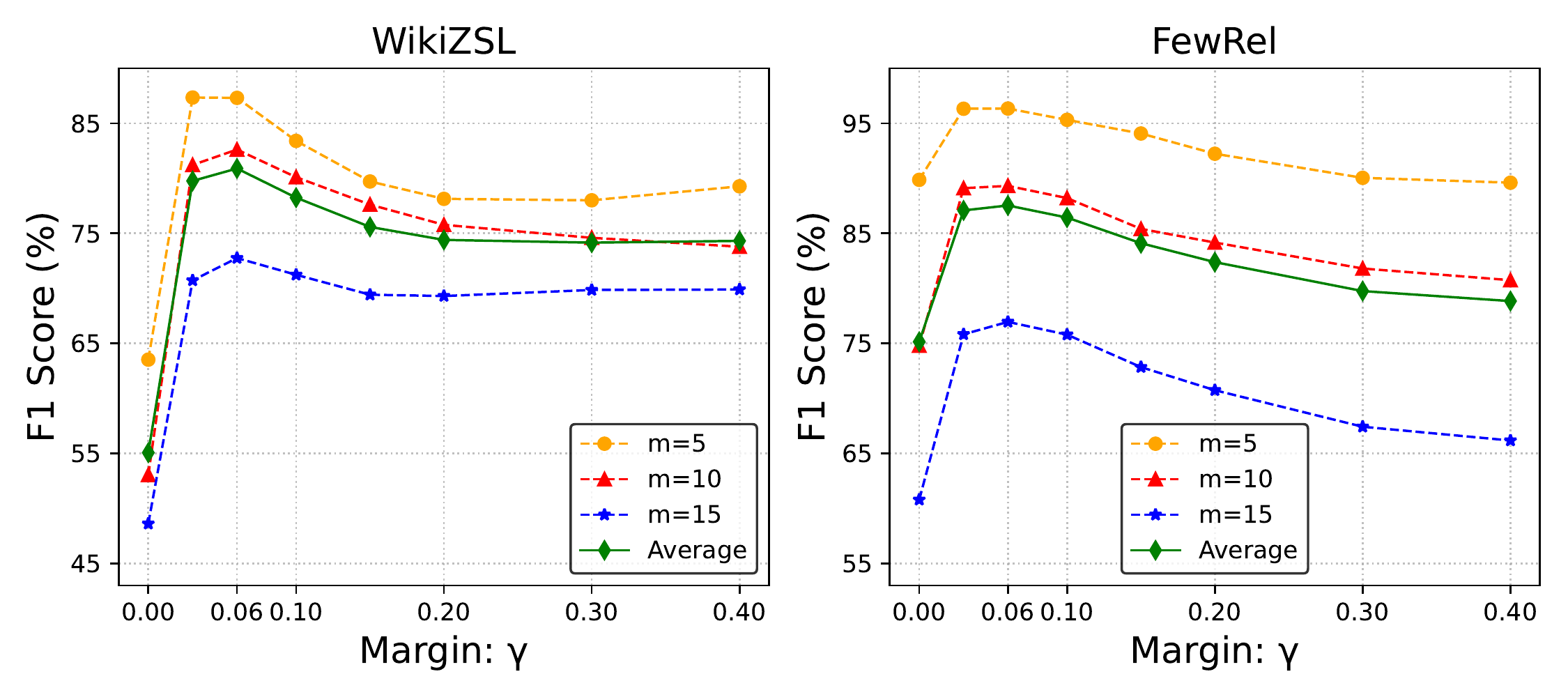}
            \caption{Model performance with different $\gamma$}
            \label{fig:hyper}
        \end{figure}
$\gamma$ is an important hyper-parameter in our optimization objective. It means that the matching score of the positive pair should be at least $\gamma$ higher than that of the negative pairs. In this subsection, we conduct experiments on two datasets (single class selection process) to study the influence of the value $\gamma$ on matching performance. From fig. \ref{fig:hyper} we can obtain the following observation. First, as $\gamma$ increases from 0, the model learns the difference between positive and negative pairs, thus assigning higher matching scores to input and correct description. When $\gamma$ increases to a critical value (i.e., 0.06), the performance begins to decline gradually. This indicates that a too-large value makes the model overfit to known relation in  training set, and then lose generalization. Finally, even if $\gamma$ is increased to a very large value, the matching does not crash. This shows that our method has good robustness.

\subsection{The Influence of Entity Description}
In a description, there are usually only one or two words that identify the entity type. As shown in tab. \ref{tab:ent}, we explore how to build a high-quality entity representation based on the words for entity matching. A simple way is to directly encode the words as entity representation. However, insufficient entity information limits matching performance. We further use synonyms to enrich entity information and improve the $F_1$ score by $3.85\%$ in Wiki-ZSL. In order to further construct a complete and fluent entity description, we fill the synonym-extended word sequence into the template slot. Compared with the original \textit{keyword} method, the two operations improve $F_1$ score by $6.78\%$ and $1.75\%$ on the two datasets respectively.
        \begin{table}
            \centering
            \resizebox{\columnwidth}{!}{
            \begin{tabular}{cl lll}
            \toprule
            \textbf{Dataset} & \textbf{Method} & Prec. & Rec. & $F_1$\\
            \midrule
            \multirow{3}{*}{\textbf{Wiki-ZSL}}
            &Keyword & $62.03$ & $59.12$ & $60.54$\\
            &Synonyms & $65.28$ & $63.53$ & $64.39$\\
            &\textbf{Template}& $67.31$ & $67.33$ & $\textbf{67.32}$\\
            \hline
            \multirow{3}{*}{\textbf{FewRel}}
            &Keyword & $72.01$ & $71.89$ & $71.91$\\
            &Synonyms & $72.24$ & $71.68$ & $71.96$\\
            &\textbf{Template}& $73.80$ & $73.52$ & $\textbf{73.66}$\\
            \bottomrule
            \end{tabular}
            }
            \caption{Comparison of different construction methods of entity descriptions (m=15).}
            \label{tab:ent}
        \end{table}  
        
\section{Conclusions}
In this work, we propose a fine-grained semantic matching method for ZeroRE. This method explicitly models the matching pattern of relational data, by decomposing the similarity score into entity and context matching scores. We explore various ways to enrich entity description and thus facilitating high-quality entity representation. The context distillation module effectively reduces the negative impact of irrelevant components on context matching. Experimental results show that our method achieves higher matching $F_1$ score and has an inference speed 10 times faster when compared with SOTA methods.

\section*{Limitations}
Elaborated relation descriptions are the foundation of the matching-based methods to achieve superior performance. Although we have proposed some ways to enrich the entity information in the descriptions, it is still a promising direction to explore more diversified and effective ways to enrich relation description (e.g. ensemble of multiple descriptions). We leave this as our future work.

\section*{Acknowledgements}
The authors wish to thank the anonymous reviewers for their helpful comments. This work was partially funded by National Natural Science Foundation of China (No.61976056,62076069,62206057), Shanghai Rising-Star Program (23QA1400200), and Natural Science Foundation of Shanghai (23ZR1403500).


\bibliography{anthology,custom}
\bibliographystyle{acl_natbib}
\end{document}